\documentclass[10pt,twocolumn,letterpaper]{article}

\usepackage[pagenumbers]{cvpr} 

\usepackage{graphics} 
\usepackage{epsfig} 
\usepackage{mathptmx} 
\usepackage{times} 
\usepackage{amsmath} 
\usepackage{amssymb}  
\usepackage[T1]{fontenc}
\usepackage{booktabs}
\usepackage{multirow}
\usepackage{graphicx} 
\usepackage{array} 
\usepackage{amsmath}
\usepackage{float}
\usepackage{placeins} 
\usepackage[most]{tcolorbox}
\usepackage{xcolor}
\usepackage{enumitem}
\usepackage{float}
\usepackage{makecell}

\definecolor{cvprblue}{rgb}{0.21,0.49,0.74}
\usepackage[pagebackref,breaklinks,colorlinks,allcolors=cvprblue]{hyperref}

\def\paperID{*****} 
\def\confName{CVPR}
\def\confYear{2025}

\title{Face-MLLM: A Large Face Perception Model}

\author{Haomiao Sun$^{1,2}$, Mingjie He$^{1,2}$, Tianheng Lian$^{1,2}$, Hu Han$^{1,2}$, Shiguang Shan$^{1,2}$\\
$^1$ Key Lab of AI Safety, Institute of Computing Technology, CAS, Beijing, 100190, China\\
$^2$ University of Chinese Academy of Sciences, Beijing, 100049, China \\
{\tt\small haomiao.sun@vipl.ict.ac.cn, \{hemingjie, liantianheng24s, hanhu, sgshan\}@ict.ac.cn}}

\begin{document}
\maketitle

\begin{abstract}
Although multimodal large language models (MLLMs) have achieved promising results on a wide range of vision-language tasks, their ability to perceive and understand human faces is rarely explored. In this work, we comprehensively evaluate existing MLLMs on face perception tasks. The quantitative results reveal that existing MLLMs struggle to handle these tasks. The primary reason is the lack of image-text datasets that contain fine-grained descriptions of human faces. To tackle this problem, we design a practical pipeline for constructing datasets, upon which we further build a novel multimodal large face perception model, namely Face-MLLM. Specifically, we re-annotate LAION-Face dataset with more detailed face captions and facial attribute labels. Besides, we re-formulate traditional face datasets using the question-answer style, which is fit for MLLMs. Together with these enriched datasets, we develop a novel three-stage MLLM training method. In the first two stages, our model learns visual-text alignment and basic visual question answering capability, respectively. In the third stage, our model learns to handle multiple specialized face perception tasks. Experimental results show that our model surpasses previous MLLMs on five famous face perception tasks. Besides, on our newly introduced zero-shot facial attribute analysis task, our Face-MLLM also presents superior performance.
\end{abstract}    
\section{Introduction}
\label{sec:Intro}

As one of the most active research fields of computer vision, the study of face perception has achieved remarkable progress. Researchers have developed a large number of advanced deep models for various face perception tasks, such as facial attribute classification \cite{miyato2018virtualadversarialtrainingregularization,noroozi2017unsupervisedlearningvisualrepresentations,2021Learning,zheng2022general}, expression analysis \cite{chang2018facialexpressionrecognitionbased, Li_2022, 2014, wang2018humanemotionalfacialexpression}, and age estimation \cite{Cao_2020,kuprashevich2023mivolomultiinputtransformerage,2015Age,tian2016jointgenderclassificationage}. Although these models have demonstrated promising results, they are still restricted to pre-defined tasks and lack the zero-shot capability of new tasks. The development of a general-purpose face perception model remains an ongoing research problem.

\begin{figure}
\centering
    \includegraphics[width=\linewidth]{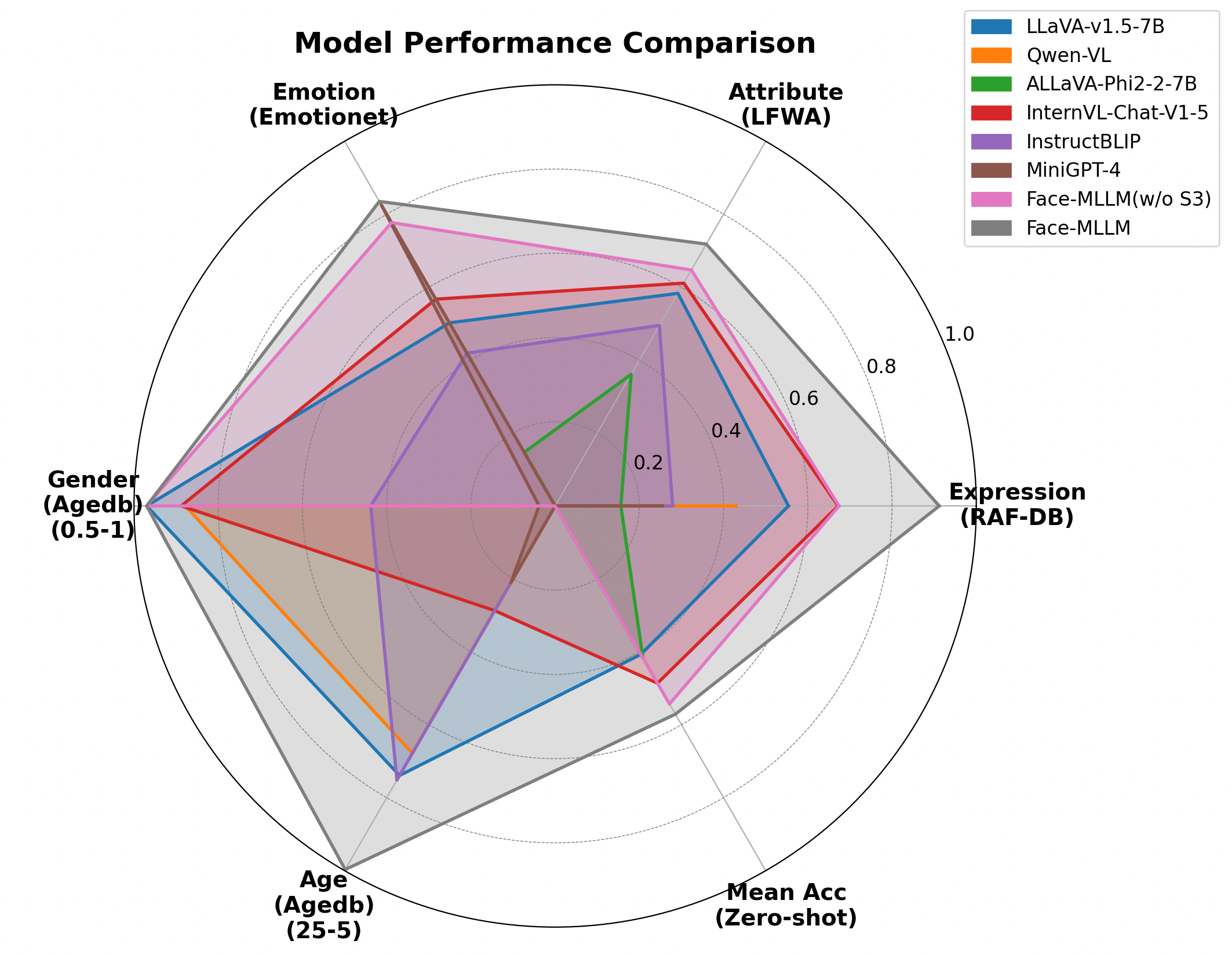}
    \caption{Our Face-MLLM model demonstrates superior performance in both traditional and zero-shot face perception tasks, showcasing its robustness and versatility in handling various face perception challenges.}
\label{fig:radar}
\end{figure}

Meanwhile, the quick evolution of the Multi-modal Large Language Model (MLLM) has shown very impressive results on diverse open-ended vision-centric tasks \cite{qwenvl,chen2024internvlscalingvisionfoundation,liu2023visualinstructiontuning,zhu2023minigpt4enhancingvisionlanguageunderstanding}. These models leverage the knowledge from natural language tasks to enhance performance in computer vision domains, leading to promising results on a wide range of vision-language tasks. Despite these advancements, their ability to perceive and understand human faces is rarely tested. 

\begin{table*}[!ht]
    \centering
    \renewcommand{\arraystretch}{1.1}
    \resizebox{\linewidth}{!}{
        \begin{tabular}{p{3cm}p{18cm}}
        \toprule[2pt]
                \multicolumn{2}{c}{\textbf{\normalsize Face Description Example}} \\
                \toprule
            \multicolumn{2}{c}{\includegraphics{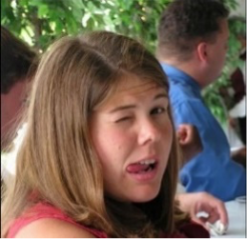}} \\
            \normalsize User prompt & \small Please describe the person in the picture in detail according to his/her face. \\
            \midrule
            \normalsize \textbf{Face-MLLM} & \small \textcolor{blue}{The person in the picture is a young adult female with fair skin and light brown hair.} \textcolor{red}{She has a round face} \textcolor{blue}{with a slightly upturned nose and a full mouth, which is open in a playful manner. Her eyes are light brown, and her eyebrows are thin and lightly arched. The expression on her face is one of mild amusement or playfulness, indicated by her tongue sticking out and her slightly raised eyebrows. The red top she is wearing is a solid color, and it appears to be a casual, possibly summer garment.} The background is blurred, suggesting that the focus is on her, and the lighting is soft, which is typical for a daytime outdoor setting. \\
            \midrule
            \normalsize Gemini-1.5-Pro & \small \textcolor{blue}{The picture shows a young woman with fair skin. She has long, straight brown hair} \textcolor{red}{parted on the side.} \textcolor{blue}{She has a playful expression, winking with her left eye while sticking out her tongue.} \textcolor{red}{She has a rounded face} \textcolor{blue}{and a subtle smile line on her right cheek.}  \\
            \midrule
            \normalsize GPT-4o & \small \textcolor{blue}{The person has shoulder-length light brown hair that is straight and smooth. They are wearing a red sleeveless top.} The background suggests an outdoor setting with greenery, and there are other people nearby. \\
            \midrule
            \normalsize LLaVA-v1.5 & \small \textcolor{blue}{The person in the picture is a woman with long}, \textcolor{red}{curly hair.} \textcolor{blue}{She is making a funny face, with her tongue sticking out} \textcolor{red}{and her eyes wide open}. She appears to be enjoying herself and is likely posing for a playful or humorous photo. \\
            \midrule
            \normalsize InternVL-v1.5 & \small \textcolor{blue}{The young woman in the photo has light brown hair and fair skin}. \textcolor{red}{She looks surprised or disgusted, holding something near her mouth}. The background suggests an outdoor social gathering with green foliage. \textcolor{blue}{She wears a red top but no more details about clothes are visible.} \\
                \midrule
                \normalsize MiniGPT-4 & \small \textcolor{blue}{The person in the picture is a woman with long brown hair and a red dress.} \textcolor{red}{She has a round face} \textcolor{blue}{and a small nose.} \textcolor{red}{Her eyes are closed} \textcolor{blue}{and her mouth is open,} \textcolor{red}{as if she is smiling. She has a necklace and earrings on, and her hands are on the table in front of her}. There are other people in the background, sitting at tables and talking. The setting appears to be a restaurant or some other type of gathering place.\\
                \midrule
                \normalsize Qwen-VL & \small \textcolor{blue}{The image shows a young woman in a red shirt} \textcolor{red}{making a silly face,} \textcolor{blue}{sticking out her tongue and winking at the camera.}\\
                \midrule
                \normalsize ALLaVA-Phi2 & \small T\textcolor{blue}{he person in the picture is a woman with a joyful expression. She has a fair complexion and her hair is blonde} \textcolor{red}{and wavy. Her eyes are blue}, and \textcolor{blue}{she is wearing a red top}. \textcolor{red}{She is puckering her lips in a playful manner, which could indicate she is smiling or laughing}. The overall impression is that of a relaxed and happy individual, possibly enjoying a social gathering.\\
                \midrule
                \normalsize InstructBLIP & \small \textcolor{blue}{The person in the picture is a young woman wearing a red dress and sitting at a dining table. She has her mouth open and appears to be making a funny face or expression.} \textcolor{red}{Her facial expression and body posture suggest that she might be surprised or amused by something happening around her.}\\
            \bottomrule[2pt]
    \end{tabular}}
    \caption{Qualitative comparison of different models. We let different models describe a given face image in detail. We use blue and red to denote the attributes that are accurately and inaccurately predicted by each model, respectively. * GPT-4o and Gemini-1.5 Pro are two well-known \textbf{closed-source} multi-modal large language models.}
 \label{table:1}
\end{table*}

To address this issue, we comprehensively evaluate existing MLLMs on face perception tasks. The quantitative results reveal that existing MLLMs struggle to handle the fine-grained facial perception tasks. This can be primarily attributed to the lack of image-text datasets containing detailed descriptions of human faces in their training data. We notice that previous researchers have invested considerable effort in collecting a large-scale image-text dataset named as LAION-Face ~\cite{zheng2022general}. With the help from a face detection model, the dataset has ensured that each image has at least one face. However, the text in LAION-Face has not undergone any screening. As a consequence, it cannot be guaranteed that the text contains descriptions of the face that corresponds with it, and that's actually the case. Furthermore, although other traditional face perception datasets, e.g. CelebA \cite{liu2015deep} and RAF-DB \cite{li2017reliable} have well-defined and manually labeled facial attribute annotations, they are not structured in the question-answer (QA) format required by MLLM. To tackle these problems, we design a practical pipeline for constructing datasets. Specifically, we employ Gemini~\cite{geminiteam2024geminifamilyhighlycapable} for automatic re-annotation of LAION-Face, enhancing it with more detailed captions and detailed facial attribute descriptions. Although Gemini’s attribute annotations contain a certain proportion of errors, we can still obtain a basically usable training set with sufficient labels via a properly designed label cleaning mechanism. Additionally, we reformulate traditional manually annotated face perception datasets into the QA format, thereby creating a large-scale dataset suitable for MLLM.

Building upon these enriched datasets, we develop a novel three-stage training method for Face-MLLM. This first stage utilizes face caption data from the re-annotated LAION-Face dataset. The primary goal is to align the visual and textual representations, creating a unified space where the model can effectively associate facial images with their corresponding textual descriptions. In the second stage, we leverage a medium-quality but large-scale dataset derived from the re-annotated LAION-Face. The extensive data allows the model to learn from a diverse range of facial structures and attributes, enabling it to develop a broad understanding of human faces. After that, the model can get a better foundational and general face perception capabilities, and can better grasp facial features, variations, and common characteristics across a wide spectrum of faces. The final stage employs medium-scale but high-quality data from traditional face perception datasets, reformatted into QA pairs. This stage aims to refine the model's performance on specific face perception tasks and improve its ability to provide structured responses to diverse face-related queries. The high-quality annotations in this stage allow for precise fine-tuning of the model's capabilities. Our experimental results demonstrate that this three-stage training approach significantly enhances the performance of Face-MLLM across various face perception tasks. Furthermore, on our newly introduced zero-shot facial attribute analysis task, Face-MLLM also outperforms existing models, showcasing its robustness and versatility in handling diverse face perception challenges. As shown in Table \ref{table:1}, our method can provide detailed and accurate descriptions of face images and obtains description results comparable to those of well-known closed-source MLLMs such as GPT4-o \cite{gpt4o} and Gemini-1.5-Pro \cite{geminiteam2024geminifamilyhighlycapable}. The main contributions of this paper are as follows:
\begin{itemize}
    \item We present a comprehensive evaluation of existing MLLM models on face perception tasks, revealing the limitations of current general-purpose models in this domain.
    \item We develop a low-cost data construction pipeline to overcome the scarcity of suitable training data, including the re-annotation of LAION-Face and the re-formulation of traditional face datasets into MLLM-compatible formats.
    \item Based on these datasets, we propose a three-stage training approach that effectively enhances the performance of Face-MLLM on both traditional and zero-shot face perception tasks.
    \item We establish a new benchmark for zero-shot facial attribute analysis, demonstrating the superior performance of Face-MLLM compared to existing state-of-the-art MLLMs.
\end{itemize}
\begin{figure*}[!t]
\centering
    \includegraphics[width=0.8\linewidth]{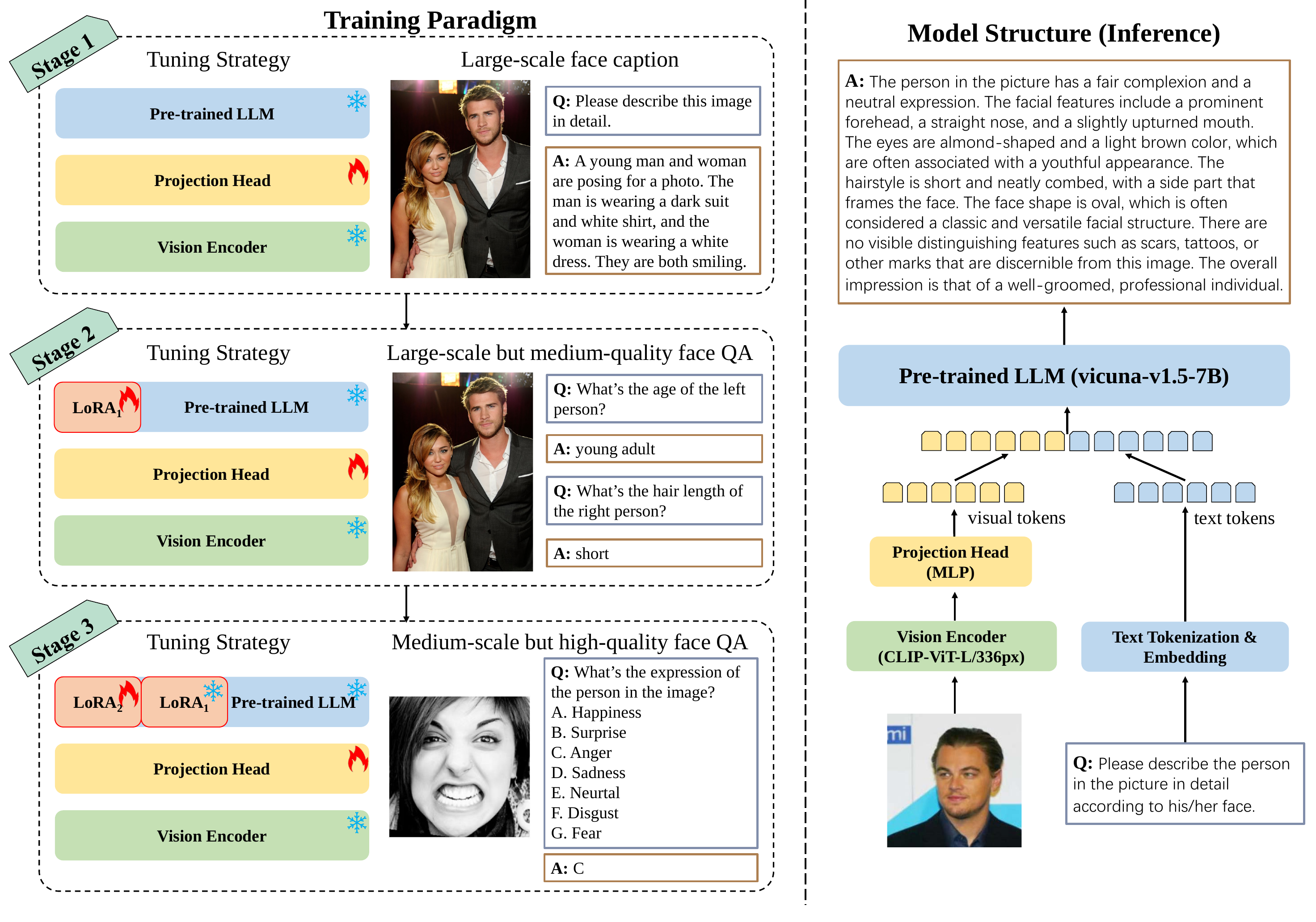}
    \caption{Training paradigm and architecture of Face-MLLM. The left side illustrates our three-stage training strategy, including representative examples of training data for each stage. The right side depicts the model's structural components, alongside an example of face description task.}
    \label{fig:stage_data}
\end{figure*}

\section{Related Work}
\label{sec:formatting}
\subsection{Face Perception}
Face perception involves the detection, analysis, and understanding of human facial features, encompassing tasks such as facial parsing~\cite{chen2016attentionscalescaleawaresemantic, jackson2016cnncascadelandmarkguided,kowalski2017deepalignmentnetworkconvolutional, te2020edgeawaregraphrepresentationlearning,2017Learning}, facial attribute recognition~\cite{miyato2018virtualadversarialtrainingregularization,noroozi2017unsupervisedlearningvisualrepresentations,2021Learning,zheng2022general}, age/gender estimation~\cite{Cao_2020,kuprashevich2023mivolomultiinputtransformerage,2015Age,tian2016jointgenderclassificationage}, head pose estimation~\cite{Cobo_2024,2020Multi,yang2015facealignmentassistedhead,zhou2020whenetrealtimefinegrainedestimation}, and facial expression recognition~\cite{chang2018facialexpressionrecognitionbased, Li_2022, 2014, wang2018humanemotionalfacialexpression}. Although promising performance has been achieved, these methods primarily focus on developing specific models for each single task and cannot handle multiple face perception tasks in a unified model. Some pioneer works, i.e., MTCNN~\cite{2016Joint}, HyperFace~\cite{ranjan2017hyperfacedeepmultitasklearning}, and AIO~\cite{ranjan2016allinoneconvolutionalneuralnetwork} attempt to build multi-task models, but the commonly used multi-task learning pipeline can only enable these models to perform a few highly correlated tasks, such as face detection and facial landmark detection. Recently, with the quick development of transformers, researchers now have a strong backbone structure with sufficient capacity for multiple tasks. FaceXFormer~\cite{narayan2024facexformerunifiedtransformerfacial} , Q-Face~\cite{sun2024taskadaptiveqface} , Faceptor~\cite{qin2024faceptorgeneralistmodelface} and SwinFace~\cite{Qin_2024} are Transformer-based face perception models. FaceXFormer~\cite{narayan2024facexformerunifiedtransformerfacial} introduces a parameter-efficient decoder called FaceX, which jointly processes facial and task tokens, thereby learning general and robust facial representations for multiple tasks. Q-Face~\cite{sun2024taskadaptiveqface} introduces a task-adaptive decoder that utilizes cross-attention for task-related feature extraction. Faceptor~\cite{qin2024faceptorgeneralistmodelface} develops the Layer-Attention mechanism to further optimize multi-task performance. SwinFace~\cite{Qin_2024} employs the Swin Transformer as its backbone and uses the MLCA (Multi-Level Channel Attention) module to address conflicts in multi-task learning. Despite these advancements, most existing methods are still restricted to pre-defined tasks and lack the zero-shot capability on new tasks. How to enable face perception models to recognize open vocabulary facial attributes remains an open research question.

\subsection{Multi-modal Large Language Models}
In recent years, the proliferation of parameters and training data has led to remarkable performance by MLLMs in visual-language tasks. Current research in this field is primarily focused on two fronts. One line of research is architectural innovation. Most of the open-source MLLMs currently follow the architecture of Vision Encoder - Vision-Language adapter - LLM. In the selection of Vision Encoders, Qwen-VL~\cite{qwenvl} employs ViT~\cite{dosovitskiy2021imageworth16x16words} as the visual encoder, leveraging pre-trained weights from Openclip. MiniGPT-4~\cite{zhu2023minigpt4enhancingvisionlanguageunderstanding} utilizes a visual encoder comprising ViT~\cite{dosovitskiy2021imageworth16x16words} integrated with a text-image alignment module, referred to as the Q-former. InternVL~\cite{chen2024internvlscalingvisionfoundation} opts for the InternViT-6B as its visual encoder, whereas LLaVA~\cite{liu2023visualinstructiontuning} relies on the CLIP ViT-L/14~\cite{radford2021learningtransferablevisualmodels} for its visual encoding needs. Regarding the choice of LLMs, Qwen-VL~\cite{qwenvl} is based on the Qwen-7B~\cite{qwen} language model. In contrast, InternVL~\cite{chen2024internvlscalingvisionfoundation} incorporates a linguistic middleware termed QLLaMA. Both MiniGPT-4~\cite{zhu2023minigpt4enhancingvisionlanguageunderstanding} and LLaVA~\cite{liu2023visualinstructiontuning} foundation their language models on Vicuna~\cite{vicuna2023}. For the Vision-Language adapters, Qwen-VL~\cite{qwenvl} incorporates a Position-aware Vision-Language Adapter, InternVL's~\cite{chen2024internvlscalingvisionfoundation} adapter is a MLP. Conversely, MiniGPT-4~\cite{zhu2023minigpt4enhancingvisionlanguageunderstanding} and LLaVA~\cite{liu2023visualinstructiontuning} utilize a simplistic linear projection layer for their adapters, facilitating the fusion of visual and language modalities. 

Another line of research is the exploration and improvement of training strategies. Qwen-VL~\cite{qwenvl} adopts a three-stage training approach: pre-training with 1.4 billion image-text pairs, multi-task pre-training with 100 million data covering seven major tasks, and instruction fine-tuning with 350,000 dialogues to enhance conversational capabilities. MiniGPT-4~\cite{zhu2023minigpt4enhancingvisionlanguageunderstanding} follows a two-stage training strategy. During the pre-training, the model is trained on a composite dataset comprising LAION~\cite{schuhmann2021laion400mopendatasetclipfiltered}, Conceptual Captions~\cite{changpinyo2021conceptual12mpushingwebscale}, and SBU~\cite{2011Im2Text} to acquire vision-language knowledge. Subsequently, the second-stage fine-tuning is performed using a meticulously curated high-quality image description dataset. LLaVA's~\cite{liu2023visualinstructiontuning} training includes contrastive pre-training for image understanding, followed by feature alignment and end-to-end weight updates. InternVL~\cite{chen2024internvlscalingvisionfoundation} starts with visual-language contrastive training, then freezes certain components while training new learnable queries and layers, and concludes with supervised fine-tuning of the connection to an existing LLM decoder.

While these models excel in general benchmarks, their performance in specific domains is suboptimal. Our experimental findings demonstrate that their performance on face perception tasks is rather limited. To address this issue, we propose a lage face percpetion model Face-MLLM, establish a novel benchmark for face perception evaluation, and curate a dataset comprising 150,000 facial images with detailed annotations to enhance the model's comprehension of facial features and improve performance in multi-attribute learning and facial recognition tasks.

\section{Collection of Training Data}

To address the limitations of existing datasets for face perception in MLLMs, we build two specialized face perception datasets, i.e., the Re-annotated LAION-Face Dataset and the Re-formulated Face Perception Datasets. They are built based on the existing face datasets and are carefully relabeled and organized to meet the training needs of MLLM. Figure \ref{fig:stage_data} shows some face image-text pairs used in each training stage. These well-prepared datasets not only support the model's training process but also provide a solid foundation for zero-shot learning and cross-task generalization. In this section, we will present the collection process of these two datasets and the contribution of each dataset to our training process.

\subsection{Re-annotated LAION-Face Dataset}
We first generate a large-scale but medium-quality face text-image paired data based on the LAION-Face dataset. LAION-Face dataset \cite{zheng2022general} is a subset of the LAION-400M dataset \cite{schuhmann2021laion400mopendatasetclipfiltered}, containing around 20 million face-caption pairs. With the help from a face detection model, the dataset has ensured that each image has at least one face. However, the text in LAION-Face has not undergone any screening. As a consequence, it cannot be guaranteed that the text contains descriptions of the face that corresponds with it, and that's actually the case. To solve this problem, we leverage the advanced vision-language model Gemini-1.0-Pro-Vision \cite{geminiteam2024geminifamilyhighlycapable} to re-annotate the faces in LAION-Face. In this way, detailed captions of 150,000 facial images are collected. These image-description pairs support the training of Stage 1. Furthermore, we generate approximately 4.75 million image-question-answer (QA) pairs based on these re-annotated face-caption pairs, which is crucial for the training of Stage 2. Our data preparation process involved the following steps:

\textbf{Re-annotation:} We leverage the advanced capabilities of Gemini-1.0-Pro-Vision to simultaneously generate detailed captions and predict facial attributes for 150,000 face images. We carefully design a detailed prompt as shown in Figure \ref{fig:prompt}. Such a prompt can guide Gemini to generate refined captions while concurrently predicting the fine-grained facial attributes for each face in the image. We also provide the Gemini model with a range of candidates for each attribute to ensure a accurate attribute annotation. After the re-annotation process, we can get rich annotations that capture a wide range of face expressions, and attributes.

\textbf{Label Cleaning:} However, the Gemini model still struggles to predict certain attributes, and in some instances, it may not provide a definitive answer. For instance, the model's ability to assess facial expressions significantly declines when faces are at extreme angles, and it struggles to accurately estimate hairstyles for individuals wearing hats. In such cases, the model may report a description such as \textit{"Cannot determine."} In response to these challenges, we clean the label generated by Gemini. Specifically, we remove ambiguous descriptions and excluded samples that lack clear facial descriptions.

While Gemini's attribute annotations still contain a certain proportion of errors, their low cost allows for large-scale labeling. By cleaning and converting these annotations into both image-caption pairs and question-answer pairs, we can create two enriched training sets. These datasets effectively support the first two stages of training and help the model to obtain a fundamental understanding of facial images. 

\begin{figure}[!t]
    \begin{tcolorbox}[title={Prompt for LAION-Face's Re-annotation}, fontupper=\scriptsize, before skip=0pt, after skip=0pt]
Suppose you are a face fine-grained attribute analyst; based on a given image, you can output both the image caption in detail and the list of fine-grained attributes for each person.

\textbf{Requirements:}
\begin{enumerate}[label=\Alph*., leftmargin=12pt]
    \item The image caption should be generated according to the fine-grained attributes.
    \item Please extract them from the image, do not imagine yourself.
    \item If there are multiple people in the image, please separate each person, point out the position of each person, and list a list of fine-grained attributes for each person.
    \item The list of fine-grained attributes should be formatted as follows:\\
    \texttt{* Attribute name: Attribute value}\\
    For example:
    \texttt{* Gender: Male\\
    * Age: Child\\
    * Hair color: Black}
    \item The fine-grained attributes include but are not limited to the following:
    \begin{enumerate}[label=\arabic*.,start=0]
        \item position
        \item age (infant, toddler, child, teenager, young adult, middle-aged, elderly)
        \item gender (male, female)
        \item race (East Asian, Southeast Asian, South Asian, Central Asian, West Asian, African, European, Native American)
        \item Hair color (black, brown, blonde, red, gray, white, etc.)
        \item Hair length (long, medium, short, bald)
        \item Hair type (straight, curly, wavy)
        \item Bangs (with bangs, without bangs)
        \item Hairline (high, low)
        \item Eye size (big eyes, small eyes)
        \item Eye Shape (Round, Almond, Phoenix)
        \item Double eyelids (double eyelids, single eyelids)
        \item Distance between eyes (wide, narrow)
        \item Eye corners (upward, downward)
        \item Bags under eyes (with bags, without bags)
        \item Dark Circles (with dark circles, without dark circles)
        \item Eye color (black, brown, blue, green, etc.)
        \item Nose size (big nose, small nose)
        \item Nose height (high bridge, low bridge)
        \item Nose width (wide nose, narrow nose)
        \item Nose tip shape (rounded tip, pointed tip)
        \item Lip thickness (thick lips, narrow lips)
        \item Lip color (red lips, pink lips)
        \item Mouth corners (upturned, downturned)
        \item Face shape (round face, square face, goose egg face, melon face, long face, diamond face)
        \item Chin shape (pointed chin, round chin, square chin)
        \item Cheekbones (high cheekbones, low cheekbones)
        \item Skin color (fair, yellowish, wheatish, tanned)
        \item Skin texture (smooth, rough)
        \item Freckles (freckled, freckle-free)
        \item Moles (with or without)
        \item Beard (bearded, unshaven)
        \item Eyeglasses (glasses, no glasses)
        \item Hat (Hat, no hat)
        \item Expression (happy, sad, angry, surprised, disgusted, fearful)
        \item Makeup (make-up, face)
        \item Jewelry (earrings, necklace, etc.)
    \end{enumerate}
\end{enumerate}
\end{tcolorbox}
\vspace{5pt}
\caption{The prompt for re-annotation of the LAION-Face data. This prompt can guide Gemini-1.0-Pro-Vision to perform both image caption and face attribute classification tasks concurrently.}
\vspace{-20pt}
\label{fig:prompt}
\end{figure}

\subsection{Re-formulated Face Perception Datasets}

Unlike the aforementioned re-annotated dataset, traditional widely-used face perception datasets are manually labeled with accurate annotations and a uniform format. In order to utilize these datasets, even though they were not originally designed for MLLM, we convert them into question-answer pairs. This re-formulated dataset complements our re-annotated LAION-Face dataset and helps to learn a more precise face perception.

We carefully chose a wide range of datasets to make our model expertise in more face perception tasks. The topics of these datasets cover facial expression recognition, age, and gender estimation, facial attribute recognition, facial action unit detection, and head pose estimation tasks. To help the model perform better on different face perception tasks, we further increase the diversity of our QA pairs. The training QA pairs have the following characteristics:
\begin{itemize}
\item \textbf{Diverse Face Perception Datasets.} To enrich the diversity of data that the model is exposed to, we also integrate divese face perception datasets for Stage 3. Specifically, we introduce UTK-Face \cite{zhifei2017cvpr}, AgeDB \cite{moschoglou2017agedb} for age estimation, AffectNet \cite{mollahosseini2017affectnet}, RAF-DB \cite{li2017reliable} for facial expression recognition, BIWI \cite{fanelli2011real} for face pose estimation, EmotioNet \cite{fabian2016emotionet} for face action unit detection, and CelebA \cite{liu2015deep}, LFWA \cite{liu2015deep} for face attribute analysis. The utilization of divese face percption datasets improves the model's performance in crucial areas of face perception. Meanwhile, it promotes a balanced distribution of data across different tasks. This approach helps prevent the model from becoming too fixated on any aspect of face perception, reducing the risk of over-fitting and enhancing the model's adaptability in diverse face perception tasks.
\item \textbf{Various Question Types}: We use many different types of questions, instead of sticking to one format. This helps the model understand and answer various types of questions, similar to real-world situations.
\item \textbf{Random Answer Choices}: To enhance the accuracy of our model, we rearrange the order of options for multiple-choice and yes/no questions. This approach encourages the model to comprehend the question thoroughly.
\item \textbf{Additional Attribute Explanations}: We include brief descriptions of specific attributes when detecting action units and estimating facial attributes. By utilizing both visual and textual information, our model is better able to perform these tasks. In addition, the introduction of descriptions of attributes during the training process further enhances the model's focus on textual information, which allows the model to perform better in zero-shot face attribute analysis tasks.
\end{itemize}

By combining these approaches, the diversity of the dataset is significantly improved, which complements the re-annotated Laion-Face dataset. The high-quality annotations refine the model’s performance on specific face perception tasks, and can guide the model to provide structured responses to diverse face-related queries.

\section{Face-MLLM}
\subsection{Network Architecture}

Face-MLLM utilizes a streamlined architecture similar to LLaVA-v1.5 \cite{liu2023visualinstructiontuning}. At the beginning, each image is encoded by CLIP-ViT \cite{radford2021learningtransferablevisualmodels}, and is mapped to the downstream LLM's token embedding space via a multi-layer perceptron (MLP). Subsequently, all tokens are fed into the LLM for auto-regressive generation. This architecture enables the model to respond to diverse instructions by capturing relationships between different image patches through the LLM, leveraging the sophisticated reasoning capabilities of large language models. With Proper training strategy, such capabilities can help model excel in diverse face perception tasks.

\subsection{Training Strategies}

As illustrated in Figure 2, our training strategy comprises three key stages and each stage builds upon its predecessor. In the first two stages, our model learns visual-text alignment and basic visual question answering capability, respectively. In the third stage, our model learns to handle multiple specialized face perception tasks.

\textbf{1) Stage 1:} The primary objective of this initial stage is to effectively align visual features with the LLM's word embedding space. We utilize a dataset comprising 150,000 captioned facial images and 660,000 image-text pairs from general scenarios. During this stage, we exclusively train the parameters of the model's MLP layer, while freezing the parameters of the vision encoder and LLM. This alignment ensures the model's ability to align the visual and textual representations, creating a unified space where the model can effectively associate facial images with their corresponding textual descriptions.

\textbf{2) Stage 2:} To obtain a basic visual question answering capability, we further train our model with a large-scale but medium-quality facial perception dataset. We convert the fine-grained attribute labels of the re-annotated LAION-Face dataset into diverse question-answer pairs, culminating in an extensive collection of 4.75 million face perception QA pairs.

A key feature of this dataset is the inclusion of positional labels, which indicate the different locations of multiple faces in the same image. This approach transforms basic questions like "\textit{Does the face in the image have attribute A?}" into more specific, position-aware queries such as "\textit{Does the leftmost face in the image have attribute A?}" By incorporating this positional awareness, we enable the model to adapt to more diverse and complex visual scenarios. This strategy not only enhances the model's ability to handle multi-face images but also improves its overall spatial reasoning capabilities in face perception tasks. Consequently, the model becomes more adept at processing and analyzing facial attributes in varied contexts.

To further enhance the model's performance in complex question-answering scenarios, we also augment the training data with 660,000 general image-text instruction tuning samples and 140,000 text-only QA pairs from ALLaVA \cite{chen2024allava}. This additional data significantly improves the model's text and image comprehension capabilities, thereby facilitating zero-shot face perception tasks. During the training process, we employ the Low-Rank Adaptation (LoRA) strategy with a rank of 16, and unfreeze the parameters in both the projection head and the large language model. This approach allows for efficient fine-tuning while preserving the model's ability to leverage pre-trained knowledge.

\textbf{3) Stage 3:} In the final stage of our training strategy, we focus on preparing the model for a diverse array of face perception tasks with optimal response capabilities. The third stage of training uses the re-formulated traditional face perception dataset, which is a higher quality manually labeled dataset. Such a medium-scale but high-quality face QA dataset greatly improves the accuracy of our model when processing face perception tasks. In addition, the model's ability to adapt to new situations and query styles is also enhanced through exposure to various problem formats, task types, and visual scenes.

To maintain the model's generalization ability and enhance the model's performance in zero-shot tasks, we also incorporate the general image-text QA dataset and text-only QA data from Stage 2. As a result, our model can comprehend complex instruction better and cope with zero-shot face perception tasks. Similar to Stage 2, we utilize the LoRA training strategy in Stage 3, but with a reduced LoRA rank of 8. This adjustment allows for more focused fine-tuning while still maintaining model efficiency. As in the previous stage, we continue to train the parameters of both the MLP and LLM components, and refine the knowledge acquired in earlier stages.

\section{Evaluation}

To comprehensively evaluate the capabilities of our model in various face perception tasks, we develop a robust benchmark. Our benchmark includes both traditional facial datasets and a novel zero-shot face attribute analysis dataset. To evaluate the model's responses to diverse face perception questions, we employ a range of quantitative metrics, including Accuracy (Acc), F1-score, and Mean Absolute Error (MAE). For classification tasks such as expression recognition and attribute detection, higher Acc and F1-score indicate superior performance. Conversely, for regression tasks like age estimation, a lower MAE is desirable, indicating more precise predictions.

\subsection{Benchmarks}
\subsubsection{Widely-used Face Perception Benchmarks}
\label{subsubsec:tfd}
To evaluate the model's performance on face perception tasks, we utilize the AgeDB \cite{moschoglou2017agedb} dataset, which comprises 16,488 images annotated with the age, name, and gender of the individuals depicted therein. Additionally, we incorporate the widely recognized RAF-DB \cite{li2017reliable} dataset, consisting of 3,068 test images that encompass seven distinct facial expressions, with each image exhibiting one of these expressions. Furthermore, we randomly select 2,000 images from the EmotioNet \cite{fabian2016emotionet} dataset for evaluation, where each image is annotated with 12 labeled Action Units (AUs). Lastly, we employ the LFWA \cite{liu2015deep} dataset, containing 6,880 images, each annotated with 40 distinct attribute labels. During the testing phase, we transform the original images and labels from various datasets into a pairwise format that pairs images with corresponding questions and answers, yielding a total of 331,000 image-question-answer pairs, as depicted in the table \ref{tab:benchmark} below. 

\begin{table}[!t]
\centering
\resizebox{\linewidth}{!}{
	\begin{tabular}{>{\scriptsize}c|c|c|>{\tiny}p{0.175\textwidth}|>{\scriptsize}c}
		\toprule[1pt]
		Dataset & Type& Image & \centering\scriptsize Question & \scriptsize Answer   \\

		\midrule
		\raisebox{-15pt}{AgeDB \cite{moschoglou2017agedb}} & Age & \raisebox{-10pt}{\includegraphics[scale=0.15]{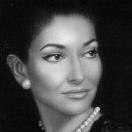}} & What is the age of the person in the picture? Estimate with a number from 1 to 100, such as 1,2,3,... & 37  \\
  
             & Gender & \raisebox{-5pt}{\includegraphics[scale=0.15]{images/agedb_12_MariaCallas_37_f.jpg}} & Is the person in the picture female? Answer directly with Yes or No. & Yes  \\
            \midrule
            \raisebox{-12pt} {RAF-DB \cite{li2017reliable}} & \raisebox{-12pt}{Expression} & \raisebox{-20pt}{\includegraphics[scale=0.05]{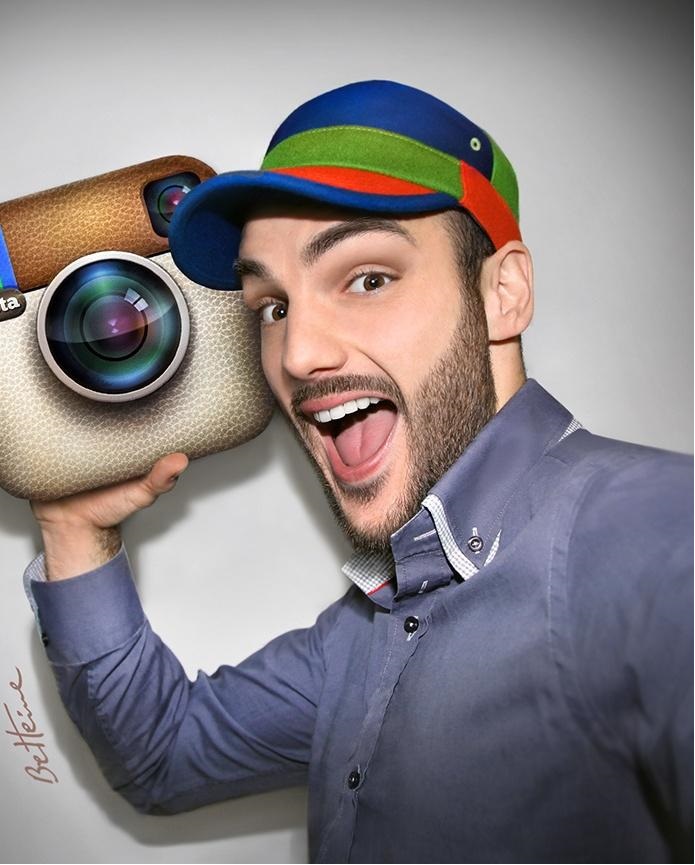}} & What's the expression of this person? A.surprise B.fear C.disgust D.happiness E.sadness F.anger G.neutral Answer with the option's letter from the given choices directly.& \raisebox{-12pt}D \\   
            \midrule

  \raisebox{-12pt}{EmotioNet \cite{fabian2016emotionet}} & \raisebox{-12pt}{Action Unit} & \raisebox{-20pt}{\includegraphics[scale=0.028]{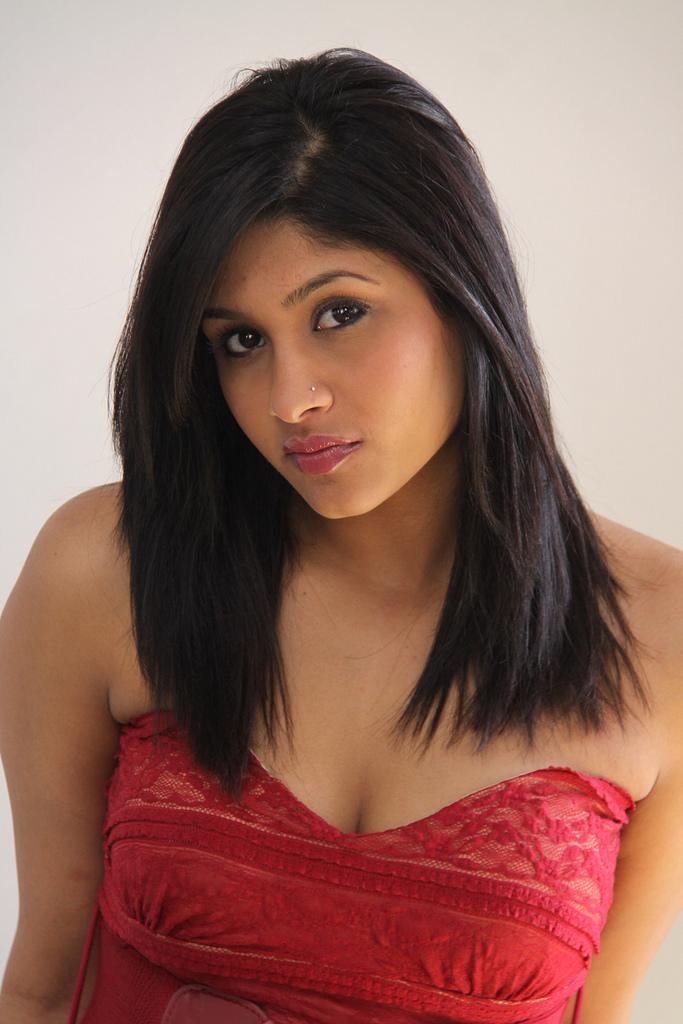}} & Inner Brow Raiser is a facial expression that involves the upward movement of the inner part of the eyebrows.Does the person in the image contains the AU of Inner Brow Raiser? Answer directly with Yes or No.  & \raisebox{-12pt}{No}   \\ 
		  \midrule
		\raisebox{-4pt}{LFWA \cite{liu2015deep}} & \raisebox{-4pt}{Attribute} & \raisebox{-10pt}{\includegraphics[scale=0.07875]{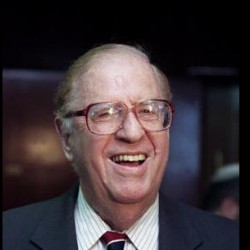}} & Does the person in the image possess the property of Bags Under Eyes? Answer directly with Yes or No. & \raisebox{-4pt}{Yes}  \\   

		\bottomrule[1pt]
\end{tabular}}
\caption{Examples for the re-formulation of face perception tasks into suitable format of MLLMs.}

\label{tab:benchmark}
\end{table}

\begin{table*}[!ht]
\centering
\renewcommand{\arraystretch}{1.1}
\resizebox{0.9\linewidth}{!}{
\begin{tabular}{c|c|c|c|c|c|c}
\toprule
\multirow{2}{*}{Model} & \multirow{2}{*}{Params} &RAF-DB \cite{li2017reliable} & LFWA \cite{liu2015deep} &  EmotioNet \cite{fabian2016emotionet} & \multicolumn{2}{c}{AgeDB \cite{moschoglou2017agedb}} \\
\cline{3-7}
&& Expression (Acc$\uparrow$) & Attribute (Acc$\uparrow$) & AU (Acc$\uparrow$)& Gender (Acc$\uparrow$) & Age (MAE$\downarrow$) \\
\midrule
MiniGPT-4 \cite{zhu2023minigpt4enhancingvisionlanguageunderstanding} &7B & 25.4 & - &  82.6  & 51.9 &  20.79\\
Qwen-VL \cite{qwenvl}&7B & 42.9 & - & - & 93.9 & 11.44 \\
InstructBLIP \cite{zhang2023instruction} &7B& 27.9 & 49.5 & 41.9 & 71.9 & 9.97 \\
ALLaVA \cite{chen2024allava}&7B & 15.6 & 36.1& 14.7 & - & - \\
LLaVA-v1.5 \cite{liu2023visualinstructiontuning}&7B & 55.4 & 58.3 & 50.2 & \textbf{98.5} & 10.22 \\
InternVL-v1.5 \cite{chen2024internvlscalingvisionfoundation}&26B & 67.2 & 61.1 & 56.7 & 94.4 & 19.26 \\
\midrule
\textbf{Face-MLLM (w/o S3)}&7B & 67.4 & 64.7 & 77.7 & \textbf{98.5} & - \\
\textbf{Face-MLLM} &7B& \textbf{91.2} & \textbf{71.8} & \textbf{83.5} & \textbf{98.5} & \textbf{5.06} \\
\bottomrule
\end{tabular}}
\caption{The comparison between Face-MLLM and other MLLMs on widely-used face perception benchmarks. We present the MAE for the age estimation task, and the classification accuracy (\%) for other tasks.}
\label{tab:example}
\end{table*}

\subsubsection{Zero-shot Face Attribute Analysis}
To evaluate the model's ability to generalize to novel facial attributes and tasks, we also collect a specialized zero-shot dataset. This dataset consists of 300 carefully selected facial images exhibiting a wide range of diverse attributes, accompanied by 760 meticulously crafted questions. We intentionally let the questions target specific facial attributes that are not included in traditional datasets, thus presenting an zero-shot face perception challenge. The facial attributes explored in this dataset include fine-grained features commonly found in Chinese traditional aesthetics:
\begin{itemize}
\item Eyelid type \cite{kiranantawat2015asian}: single eyelids, and double eyelids
\item Eye shape \cite{wang2023eyes}: phoenix eyes, almond eyes, and peach blossom eyes
\item Nose shape: upturned nose \cite{jackson2004midline}, aquiline nose \cite{wikipedia_aquiline_nose}, and low bridge nose \cite{moon2018surgical}
\item Lip shape: cherry lips \cite{butterworth2011cherry} and thick lips \cite{deQueirozHernandez2023}
\end{itemize}

Each facial image in our dataset is annotated with one major category of facial features, i.e., eyelid type, eye shape, nose shape, or lip shape. We transform each annotation into 2 or 3 distinct questions. This approach allows us to probe different aspects of the model's understanding of the same facial feature. Specifically, each generated question includes a detailed description of the new features, instructions for the model on what to do, and guidance on how the model should formulate its response. By converting each single-category annotation into multiple questions, we create a more robust and comprehensive zero-shot evaluation framework. This approach not only assesses the model's ability to recognize broad categories of facial features but also its capacity to discern and articulate subtle variations within these categories.

\begin{figure}[!t]
    \centering
    \includegraphics[width=0.95\linewidth]{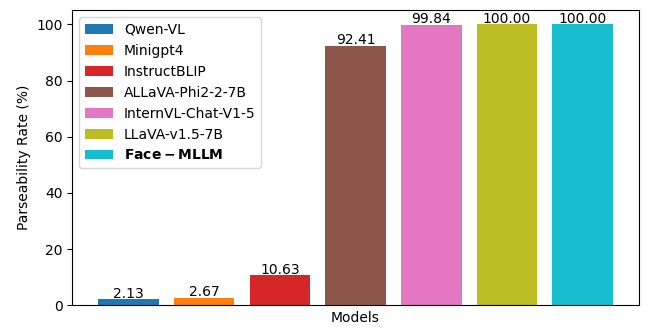}
    \caption{The probability (\%) that different models can provide properly formatted responses.}
    \label{fig:parse}
\end{figure}

\subsection{Results on Widely-used Face Perception Benchmarks}
Table \ref{tab:example} presents a comprehensive performance comparison between our Face-MLLM model and state-of-the-art open-source multi-modal large language models (MLLMs) across several common face perception tasks. This comparison provides a clear overview of our model's capabilities in relation to other leading approaches in the field.

Our Face-MLLM demonstrates exceptional performance across a spectrum of face perception tasks. In face expression recognition, attribute recognition, and action unit detection tasks, our model consistently outperforms the baseline LLaVA-v1.5, with accuracy improvements of 35.8\%, 13.5\%, and 33.3\% respectively. On AgeDB dataset, while all models show near-ceiling performance in gender prediction, Face-MLLM excels in age estimation with a MAE of 5.06, significantly lower than LLAVA-v1.5's 10.22 MAE.

Fig. \ref{fig:parse} compares the probabilities of correct or parseable format predictions across various models. The seven models are sorted from lowest to highest probability and are distinguished by different colors. It is worth noting that models like Qwen-VL, MiniGPT-4, and InstructBlip have a poor understanding of the instructions of the face perception tasks, and often fail to accurately solve the given problem. In contrast, our Face-MLLM shows an superior ability to provide reasonable responses and can accurately distinguish between different facial attributes. This is further exemplified by our use of high-precision human annotated data in stage 3, which effectively refines the model's output structure and further improves model performance.

\begin{table}[!t]
\centering
\renewcommand{\arraystretch}{1.3}
\resizebox{\linewidth}{!}{
	\begin{tabular}{c|c|c|c|c|c}
		\toprule[1pt]
		Model & \makecell{Eyelid\\type} & \makecell{Eye\\shape} & \makecell{Nose\\shape} & \makecell{Lip\\shape} & \makecell{Mean\\accuracy} \\
		\midrule
		

		LLaVA-v1.5-7B \cite{liu2023visualinstructiontuning} & 50.7 & 31.3 & 34.0 & 46.7 & 40.7\\
            ALLaVA-Phi2-2-7B \cite{chen2024allava} & 50.0 & 33.3 & 33.3 & 50.0 & 41.7\\

            MiniGPT-4-7B\cite{zhu2023minigpt4enhancingvisionlanguageunderstanding} &  50.7 &34.6 & 42.5 & 47.5& 43.8\\
            
            InternVL-V1.5-26B \cite{chen2024internvlscalingvisionfoundation} & 58.6 & 40.3 & 53.0 & 42.5 & 48.6 \\

        \midrule
            \textbf{Face-MLLM (w/o S3)} & 51.4 & 49.0 & 63.5 & \textbf{53.3} & 54.3\\   
  
		\textbf{Face-MLLM} & \textbf{59.3} & \textbf{53.9}  & \textbf{65.3} & 50.0 & \textbf{57.1}\\ 
  
		\bottomrule[1pt]
\end{tabular}}
\caption{The performance (\%) of Face-MLLM and other MLLMs on zero-shot face attribute analysis tasks.}
\label{tab:zero_shot}
\end{table}

\subsection{Results on Zero-shot Face Attribute Analysis}
Building upon the impressive performance demonstrated in Table \ref{tab:example}, we further evaluate our model's capabilities in zero-shot scenarios. As presented in Table \ref{tab:zero_shot}, we compare Face-MLLM with other state-of-the-art MLLMs, including LLaVA-v1.5, Qwen-VL, ALLaVA, InternVL-1.5, and InstructBLIP. In this zero-shot protocol, Face-MLLM has exhibited remarkable proficiency across various facial feature recognition tasks, further solidifying its superiority over the baseline model, LLaVA-v1.5. Compared to baseline, the classification accuracies are improved by 8.6\%, 22.6\%, and 31.3\% for eyelids, eyes, and noses, respectively. These significant improvements in recognizing subtle facial attributes without task-specific training demonstrate the model's generalization capabilities and deep understanding of facial structures. This ability to perform well in zero-shot tasks is particularly valuable in real-world applications where models often encounter novel or unseen instructions.

\begin{table}[!ht]
\centering
\renewcommand{\arraystretch}{1.1}
\resizebox{\linewidth}{!}{
\begin{tabular}{c|cccccccc}
\toprule
Model  & OCR & Math & Spat & Rec & Know & Gen & Overall \\
\midrule
Minigpt4 \cite{zhu2023minigpt4enhancingvisionlanguageunderstanding}  & 7.1 & 7.3 & 9.6 & 12.2 & 9.2 & 8.0 & 10.5 \\
Qwen-VL \cite{qwenvl}  & 7.4 & 0.0 & 3.9 & 16.5 & 18.6 & 18.1 & 13.0 \\
InstructBLIP \cite{zhang2023instruction}  & 22.5 & \underline{11.5} & 23.5 & 39.3 & 24.3 & 23.6 & 33.1 \\
ALLaVA \cite{chen2024allava}  & - & - & - & - & - & - & 32.2 \\
LLaVA-v1.5 \cite{liu2023visualinstructiontuning} & 26.7 & 7.7 & 25.6 & \textbf{44.9} & 22.9 & 21.5 & 32.9 \\
\midrule
\textbf{Face-MLLM (w/o S3)}  & \underline{27.2} & 6.9 & \underline{30.7} & 39.0 & \underline{32.1} & \underline{31.4} & \underline{35.4} \\
\textbf{Face-MLLM}  & \textbf{33.5} & \textbf{13.5} & \textbf{37.2} & \underline{40.7} & \textbf{32.4} & \textbf{36.5} & \textbf{37.8} \\
\bottomrule
\end{tabular}}
\caption{Performance comparison of different models on MM-Vet \cite{yu2023mm} benchmark, which provides valuable insights into the general task understanding capabilities of various multi-modal large language models. All the numbers are presented in \%.
}
\label{tab:mm-vet-comparison}

\end{table}

\subsection{Results on General Image Analysis Benchmarks}

In addition to face perception, we also evaluate our model's performance on general tasks. We conduct extensive tests using the common image-text question answering dataset MM-Vet (Multimodal-Vet) \cite{yu2023mm}. MM-Vet is a benchmark dataset designed to evaluate the general capabilities of large multi-modal language models (MLLMs). It assesses models across various domains, including OCR, mathematical reasoning, spatial understanding, visual recognition, knowledge application, and generation tasks. The results on the MM-Vet dataset can provide valuable insights into the general task understanding capabilities of various multi-modal large language models. 

Table \ref{tab:mm-vet-comparison} shows the comparison results between Face-MLLM and other SOTA MLLMs on the MM-Vet benchmark. Our Face-MLLM model shows impressive performance across a diverse range of tasks, achieving an overall score of 37.8. This performance shows Face-MLLM's strong abilities in optical character recognition (OCR), spatial reasoning, general knowledge, and knowledge-based question answering. Moreover, Face-MLLM shows significant improvements over its predecessor, Face-MLLM (without S3), across all tasks, particularly in OCR and spatial reasoning. These results demonstrate that the improvement in facial understanding has not led to a compromise in general image comprehension. This overall ability in both specific and general tasks highlights the adaptability of Face-MLLM and shows its potential for a wide range of practical applications. 

\section{Conclusion}

To overcome the limitations of current MLLMs in handling fine-grained facial analysis tasks, we develop a novel multimodal large face perception model Face-MLLM. Specifically, we first develop a low-cost data construction pipeline to overcome the scarcity of suitable training data. Meanwhile, we design a three-stage training strategy to progressively improve Face-MLLM's capabilities in visual-text alignment, basic visual question answering, and specialized face perception tasks. Experimental results show that our model surpasses previous MLLMs on a wide range of face perception tasks. It is worth noting that these improvements have not led to a compromise in general image comprehension. In addition, leveraging the inference capabilities of the large language model, Face-MLLM also demonstrates its potential in the newly introduced task of zero-shot facial attribute analysis.

\newpage
\clearpage


{
    \small
    \bibliographystyle{ieeenat_fullname}
    \bibliography{main}
}


\end{document}